\documentclass[lettersize,journal]{IEEEtran}
\usepackage{amsmath,amsfonts}
\usepackage{algorithmic}
\usepackage{algorithm}
\usepackage{array}
\usepackage[caption=false,font=normalsize,labelfont=sf,textfont=sf]{subfig}
\usepackage{textcomp}
\usepackage{stfloats}
\usepackage{url}
\usepackage{verbatim}
\usepackage{graphicx}
\usepackage{cite}
\usepackage[colorlinks,linkcolor=blue,anchorcolor=blue, citecolor=blue]{hyperref}
\usepackage{xcolor}
\graphicspath{{Figure/}}
\usepackage{booktabs}
\usepackage{amsmath}
\usepackage{microtype}
\usepackage{algorithm} 
\usepackage{algorithmic}

\begin{document}
	
	\title{Simultaneous Automatic Picking and Manual Picking Refinement for First-Break}
	
	\author{Haowen Bai, Zixiang Zhao,~\IEEEmembership{Member,~IEEE}, Jiangshe Zhang, Yukun Cui, \\Chunxia Zhang,~\IEEEmembership{Member,~IEEE}, Zhenbo Guo, Yongjun Wang
		
		\thanks{This work is supported by Key technologies for coordination and interoperation of power distribution service resource under Grant 2021YFB2401300, National Natural Science Foundation of China under Grant 12371512, Science \& Technology Research and Development Project of CNPC under Grant 2021ZG03. \textit{(Corresponding author: Jiangshe Zhang, Chunxia Zhang).}
		}
		\thanks{
			Haowen Bai, Zixiang Zhao, Jiangshe
			Zhang, Yukun Cui, Chunxia Zhang are with the School of Mathematics and Statistics, Xi’an Jiaotong University, Xi’an, Shaanxi 710049, China (E-mail:
			\{hwbaii, zixiangzhao, cuiyukun\}@stu.xjtu.edu.cn, \{jszhang, cxzhang\}@mail.xjtu.edu.cn).}
		\thanks{Zhenbo Guo is with the Geophysical Technology Research Center of Bureau
			of Geophysical Prospecting, Zhuozhou, Hebei, 072751, P.R.China (E-mail: guozhenbo01@cnpc.com.cn).}
		\thanks{Yongjun Wang is with School of Artificial Intelligence, Wenzhou Polytechnic, Wenzhou, Zhejiang, China  (E-mail: wangyjmcvti@qq.com).}
	}
	
	\maketitle
	
	\begin{abstract}
		First-break picking is a pivotal procedure in processing microseismic data for geophysics and resource exploration.
		Recent advancements in deep learning have catalyzed the evolution of automated methods for identifying first-break. 
		Nevertheless, the complexity of seismic data acquisition and the requirement for detailed, expert-driven labeling often result in outliers and potential mislabeling within manually labeled datasets.
		These issues can negatively affect the training of neural networks, necessitating algorithms that handle outliers or mislabeled data effectively.	
		We introduce the Simultaneous Picking and Refinement (\textit{SPR}) algorithm, designed to handle datasets plagued by outlier samples or even noisy labels.
		Unlike conventional approaches that regard manual picks as ground truth, our method treats the true first-break as a latent variable within a probabilistic model that includes a first-break labeling prior. 
		SPR aims to uncover this variable, enabling dynamic adjustments and improved accuracy across the dataset.
		This strategy mitigates the impact of outliers or inaccuracies in manual labels.
		Intra-site picking experiments and cross-site generalization experiments on publicly available data confirm our method's performance in identifying first-break and its generalization across different sites.
		Additionally, our investigations into noisy signals and labels underscore SPR's resilience to both types of noise and its capability to refine misaligned manual annotations.	
		Moreover, the flexibility of SPR,  not being limited to any single network architecture, enhances its adaptability across various deep learning-based picking methods.
		Focusing on learning from data that may contain outliers or partial inaccuracies, SPR provides a robust solution to some of the principal obstacles in automatic first-break picking.
		
	\end{abstract}
	
	
	\section{Introduction}
	\IEEEPARstart{F}{irst-break} picking consists of identifying and recording the first arrival time of seismic energy at each receiver after a seismic event.
	This process is fundamental to seismic data analysis, offering insights into the locational aspects of seismic events and enabling the derivation of the seismic velocity structure within the Earth's subsurface.
	Its pivotal role extends across a spectrum of geophysical explorations, encompassing both passive and active seismic methodologies.
	Notable applications include the monitoring of carbon capture and storage (CCS)~\cite{verdon2011linking,stork2015microseismic}, underground hydrogen storage (UHS)~\cite{caglayan2020technical}, early warning of earthquakes, large-scale monitoring through distributed acoustic sensing (DAS)~\cite{mestayer2011field,williams2019distributed}, modeling of seismic velocities, analyzing geologic structures, and correction of the source/receiver coordinates.
	Given the extensive applications and the growing volume and complexity of seismic datasets, the pursuit of enhancing automatic first-break picking algorithms is increasingly pressing in the realm of seismic signal processing. 
	These automatic algorithms are gradually replacing traditional manual picking, previously reliant on the expertise of seismologists.
	The transition is propelled by the advantages of automation, including efficiency, consistency, and accuracy. These qualities are becoming increasingly indispensable in the era of big data and with the advancement of high-resolution seismic imaging technologies.
	
	Early traditional automatic techniques are characterized by the employment of statistical and signal processing algorithms such as energy-based methods~\cite{stevenson1976microearthquakes,allen1978automatic,qin2021first}, correlation-based methods and Akaike information criterion (AIC)-based methods.
	Short-term average/long-term average (STA/LTA) method~\cite{stevenson1976microearthquakes,allen1978automatic,allen1982automatic,baer1987automatic,earle1994characterization} and its derivatives~\cite{li2019identification,zhang2018sta,chen2006multi,longjun2021easy,jiang2020automated,mborahimproving,han2010microseismic,lee2017improved,chen2005multi}, as prominent energy-based methods, rely on the variation of energy ratio around the sample point to determine the time of first-break.
	Notably, utilizing multi-window energy ratios enables a more robust detection of first-break~\cite{chen2006multi,kim2023first,lee2017improved,chen2005multi}.
	Energy-based methods are straightforward and computationally inexpensive, and are less effective under low signal-to-noise conditions, with their performance heavily contingent on user-defined parameters.
	Correlation-based methods~\cite{molyneux1999first,raymer2008semiautomated,senkaya2014semi}, seek to determine seismic events by correlating waveforms across receivers, offering enhanced noise resilience and improving consistency across multiple sites. These methods signify a departure from the reliance on energy metrics, aligning detection with waveform correlation.
	Akaike Information Criterion (AIC)-based methods~\cite{tozzi2020stress,leonard1999multi,maeda1985method,long2019fast} distinguish themselves by their capacity to identify the global minimum AIC value without an exclusive dependence on amplitude thresholds or waveform similarity. This attribute is particularly advantageous under challenging signal-to-noise ratios, as the AIC method pivots on statistical feature variations~\cite{leonard2000comparison}. Nevertheless, the computational demand of AIC-based methods is non-negligible~\cite{long2019fast}, and they require additional localization and calibration to ensure the accuracy of first-break picking~\cite{mborah2018enhancing}.
	
	Recently, the combination of deep learning and first-break picking has revolutionized the field of seismology~\cite{hollander2018using,chen2019automatic,zhang2020convolutional,zhu2019phasenet,han2021first,tsai2018first}, where neural networks are capable of extracting and interpreting complex features of seismic data, achieving efficient and highly accurate identification of first-break of seismic signals.
	Deep learning-based automatic picking methods are categorized into single-trace processing methods~\cite{mccormack1993first,qu2021deep,guo2020aenet} and multi-trace processing methods~\cite{hu2019first,hu2018u,sheng2022arrival,yuan2020robust}.
	Single-trace processing methods employ a strategy of extracting signals from individual traces near the sampling point and use the classification outcomes to ascertain the presence of a first-break at the center of the region.
	Within this domain, AENet~\cite{guo2020aenet} leverages convolutional neural networks (CNNs) with DBSCAN clustering analysis to retrieve the precise first-break.
	Conversely, multi-trace processing methods process the time-series signals from contiguous receivers into 2-D representations for collective analysis~\cite{jiang2023seismic,DBLP:journals/lgrs/LiuCWLG22}. 
	These methods, such as UNet~\cite{hu2019first}, SegNet~\cite{yuan2022segnet}, ResUnet~\cite{zwartjes2022first}, MSNet~\cite{sheng2022arrival}, evaluate each sampling point within the context of its neighbors, thereby harnessing the spatial correlation inherent between adjacent traces~\cite{xie2019first,zhang2020first,yuan2018seismic,ma2022intelligent,mousa2011new}. This approach tends to yield more reliable outcomes compared to single-trace picking methodologies.
	The adoption of deep learning for automatic first-break picking augments both the accuracy and efficiency of seismic data analysis. This advancement underscores the transformative potential of seismic detection, resource exploration, and broader geophysical research.
	
	Despite these advances, there are still challenges in deep learning-based first-break picking methods.
	The production of a substantial and accurately labeled training dataset is a pivotal requirement for the effective training of neural networks.
	Unfortunately, the construction of such datasets inevitably results in outlier samples and mislabeling, particularly in situations with large volumes of seismic data and intensive manual labor~\cite{qin2021method,qin2021first}.
	Outlier samples, although accurate, are like mislabeled data in that their extreme values can seriously impair the model's learning process and accuracy.
	The detrimental impact of outlier samples and label noise in the data on the performance of neural networks can be overwhelming, with numerous studies endeavoring to devise strategies that mitigate these effects and enable learning from noisy labels~\cite{DBLP:journals/corr/abs-2007-08199}. 
	Approaches such as directed graphical models~\cite{DBLP:conf/cvpr/XiaoXYHW15}, robust loss functions~\cite{DBLP:conf/cvpr/PatriniRMNQ17}, conditional random fields (CRFs)~\cite{DBLP:conf/nips/Vahdat17}, knowledge graphs~\cite{DBLP:conf/iccv/LiYSCLL17}, and neural networks~\cite{DBLP:conf/cvpr/VeitACKGB17,DBLP:conf/cvpr/WangLMBZSX18} have been employed to address label noise. These methods are particularly effective at the image level.
	This challenge also extends to pixel-level labels~\cite{DBLP:conf/eccv/YuLZFRKK18} and textual data~\cite{DBLP:conf/aaai/ZhengAD21}.
	In this paper, we propose an innovative first-break picking algorithm capable of identifying outliers or noisy labels in the data.  
    We model true first-break as latent variables within a probabilistic framework. This shifts our neural network's focus from mere manual picking to dynamically refined, potential true first-break through iterative updates.
	This method mirrors strategies previously employed in seismic data analysis. Specifically, it uses auxiliary variables in Full Waveform Inversion (FWI)~\cite{doi:10.1137/17M111328X,10.1093/gji/ggz189,8873570} to broaden the search space for viable solutions and divide the objective function into several sub-tasks.
	These sub-tasks are iteratively optimized, enhancing the model's resilience and adaptability. Similarly, Seismic Blind High-Resolution Inversion (BHRI) uses auxiliary variables to reduce dependence on initial estimates~\cite{9497758}.
	Furthermore, modeling uncertainties in deep learning forecasts with slack variables introduces these uncertainties as soft constraints in the FWI paradigm~\cite{doi:10.1190/segam2020-3417568.1}. This highlights the synergy between traditional seismic imaging methods and deep neural networks' pattern recognition capabilities.
	By modeling the true first-break as latent variables and continuously updating them, our approach not only mitigates the influence of outlier data on the learning trajectory but also corrects mislabeling through a systematic refinement process.
	
	Our main contributions are briefly summarized as follows:
	
	(1) We propose an automatic picking method capable of mining the distribution of first-break. By constructing a probabilistic model that includes potential first-break, our method resists the effects of outlier samples and mislabeling, thereby significantly improving picking accuracy.
	
	(2) The developed probabilistic model features two inference methods, capable of performing automatic first-break picking and correcting potential errors in manual picking, respectively.
	
	(3) Extensive experiments containing picking experiments, generalization experiments, noisy signal experiments, and noisy labeling experiments demonstrate the excellent performance of our method. In addition, the proposed algorithm is not limited to a specific network structure and can be generalized to arbitrary neural networks.
	
	The arrangement of the subsequent paper is as follows, in Sec. II, we describe the training process and the inference methodology of the proposed algorithm, with extensive experiments in Sec. III. Conclusions are stated in Sec. IV.
	
	\section{method}
	\begin{figure}[!]
		\centering
		\includegraphics[width=\linewidth]{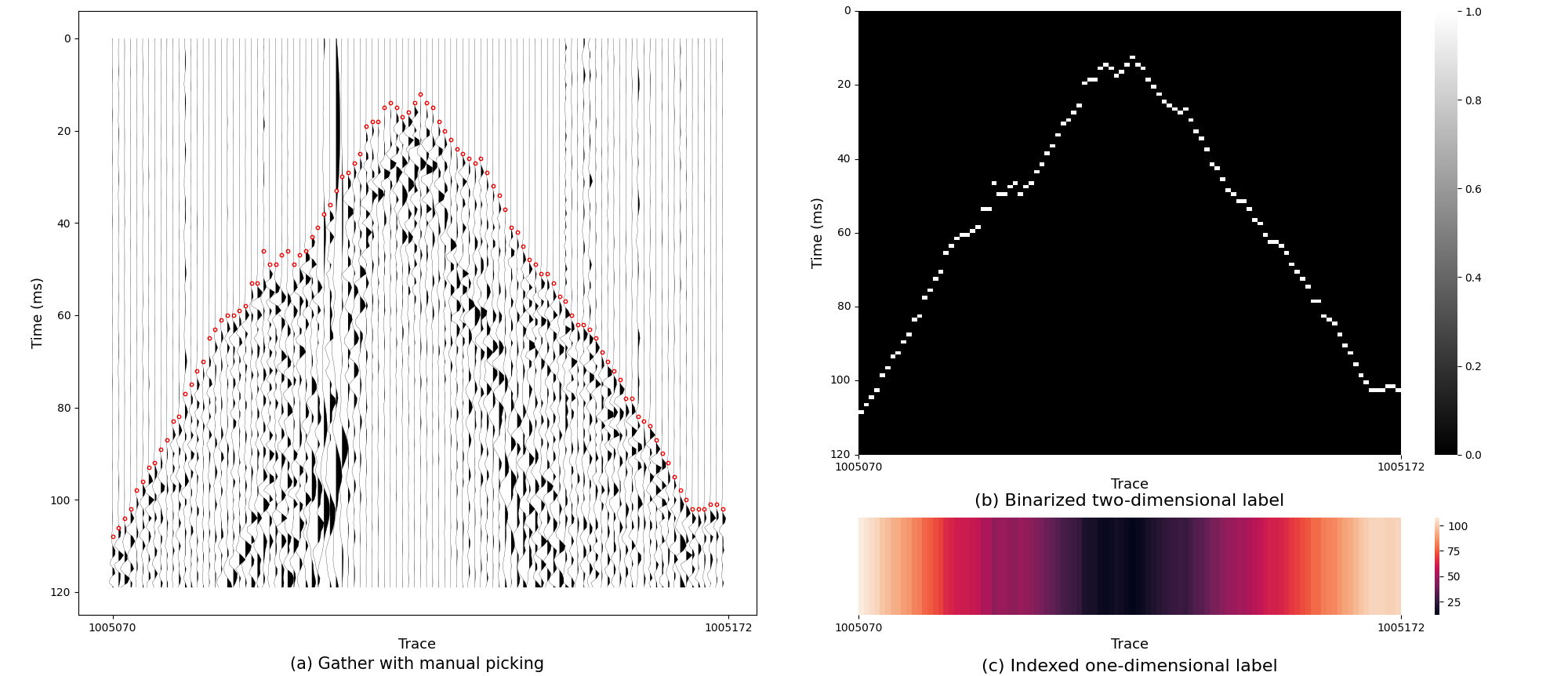}
		\caption{(a) Amplitude signals. Manual picking are marked with red circles. (b) Binarized 2-D label, the position of the first-break is 1 and the rest is 0. (c) Indexed 1-D label, the index of the first-break among all the sampling points.}
		\label{fig:label}
	\end{figure}
	Before describing the method, we need to unify the notations. In this paper, we adopt multi-trace processing for first-break picking. $x\in R^{M\times N}$ is the amplitude data received at the signal detection point, where $N$ and $M$ represent the number of traces and sampling points in each trace, respectively.
	Two forms of first-break labels are used in our analysis, as illustrated in Fig.~\ref{fig:label}. One is an indexed one-dimensional label, denoted by $t\in R^{1\times N}$. $t_k$, a number, represents the index of the first-break of $k$th trace among all the sampling points, as shown in Fig.~\ref{fig:label} (c). $y\in R^{M\times N}$ denotes the binarized 2-D label, where $k$th trace's label $y_k$ is vectorized to match the size of its corresponding signal vector. The position of the first-break is marked as 1, and the remaining elements are set to 0, as shown in Fig.~\ref{fig:label} (b).
	These two forms of labels are inherently interconvertible and maintain a direct correspondence with each other. In the rest of this paper, they will be referred to as the 1-D label $t$ and the 2-D label $y$ to streamline the discussion and analysis.
	\begin{figure*}[!]
		\centering
		\includegraphics[width=\linewidth]{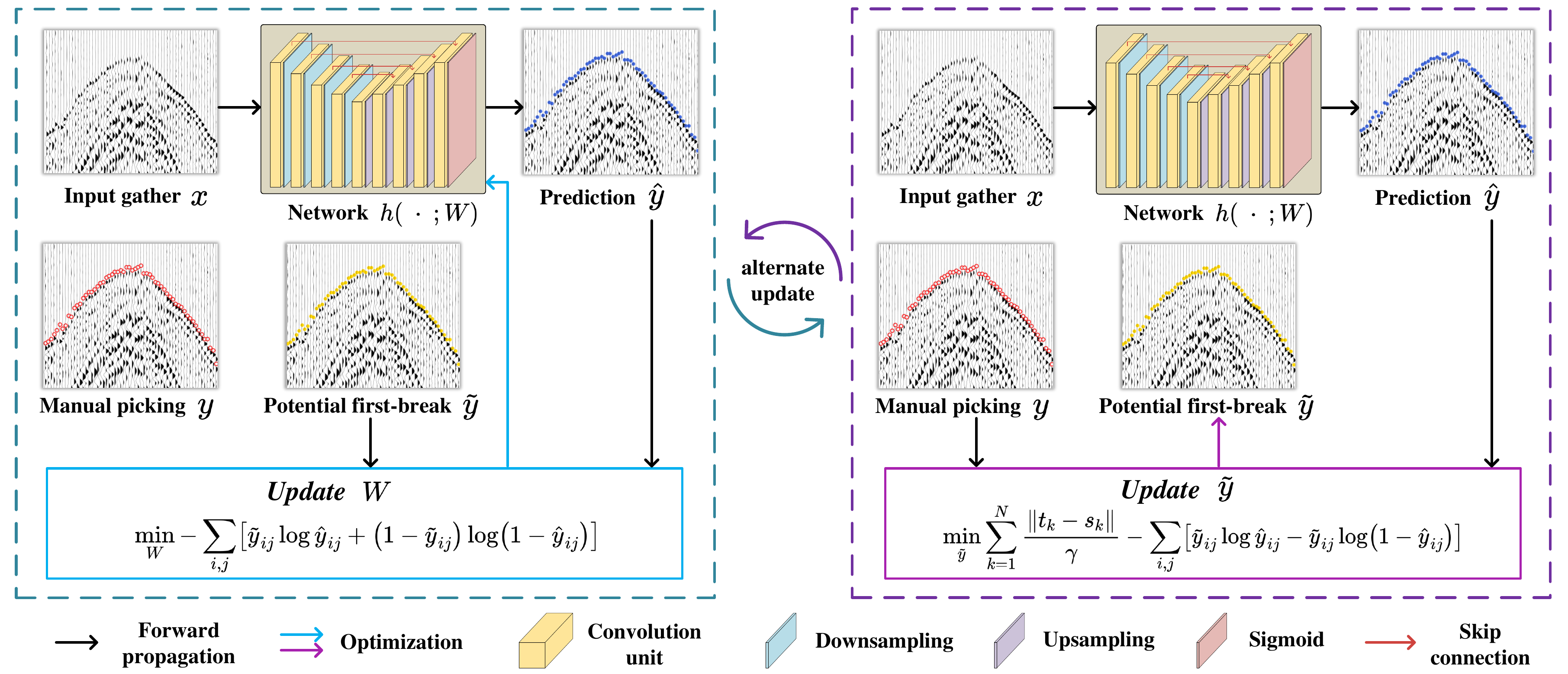}
		\caption{The training process of SPR. The network parameters $W$ and potential first-break $\widetilde{y}$ are alternately updated to simultaneously learn picking as well as manually picking refinement.}
		\label{fig:SPR}
	\end{figure*}
	
	\subsection{Probabilistic Model}
	In classification problems, the training objective of the model is usually to maximize the likelihood function:
	\begin{equation}
		\label{equ1}
		\max _{W} \mathcal{L}(W)=P(y \mid x ; W),
	\end{equation}
	where $y$, $x$, and $W$ are the labels, input gathers, and the model parameters, respectively. The same is true for many semantic segmentation-based picking methods. However, due to a variety of factors, some outlier samples or even mislabeling are inevitable, which can seriously affect the training process of neural networks. 
	In this work, we  introduce a latent first-break $\widetilde{y}$, which is different from manual picking in some traces. 
	$\widetilde{y}$ accounts for outlier sample substitution and mislabeling correction, enhancing modeling quality and reliability. 
	The model is optimized by maximizing the likelihood of the manual picking $y$ and the potential first-break  $\widetilde{y}$ as follows 
	\begin{equation}
	\begin{aligned}
		\label{equ2}
		\max _{\widetilde{y}; W} \mathcal{L}(\widetilde{y}; W)&= P(y, \widetilde{y}  \mid  x ; W)\\
		&= P(y  \mid  \widetilde{y}, x ; W) P(\widetilde{y}  \mid  x ; W)\\
		&=P(y  \mid  \widetilde{y}) P(\widetilde{y}  \mid  x ; W).
	\end{aligned}
	\end{equation}
	In the equation above, the first term represents a prior probability model for first-break point labeling. 
	This term calculates the probability of a point being labeled as $y$ given the potential first-break $\widetilde{y}$.
	We postulate that the probability of mislabeling is independent of the input $x$ and inherently unrelated to the network parameters $W$. Thus, we simplify the first term to $P(y \mid \widetilde{y})$, reinforcing our model's assumptions and its structural integrity.
	The latter item takes $\widetilde{y}$ as the prediction target of the network.
	
	Labeling misalignments and outlier samples are sparse, occurring in only a small number of traces. Thus, the Laplace distribution, known for its peaked and heavy-tailed nature, is better suited to capturing the distribution of sparse data.
	We use the Laplace distribution to model the first-break labeling prior model, i.e.
	\begin{equation}
		\label{equ3}
		P(y \mid \widetilde{y}) \propto \prod \limits_{k=1}^N \exp \left(-\frac{\left\|t_k-s_k\right\|}{\gamma}\right),
	\end{equation}
	where $s$ is the 1-D labeling of $\widetilde{y}$ as previously described.
	$t_k$ and $s_k$ are the positions of the manual picking and potential first-break of the $k$th trace, respectively. $\gamma$ represents the sensitivity parameter that controls labeling bias, with its choice depending on the specific situation. The equation suggests that each trace's first-break labels follow a Laplace distribution centered around their potential true first-break.
	
	For the prediction target, we assume a joint distribution across all samples, with each sample's probability modeled by a binary Bernoulli distribution:
	\begin{equation}
		\label{equ4}
		\begin{split}
			P\left(\widetilde{y} \mid x ; W\right) & =\prod_{i,j} P\left(\widetilde{y}_{ij} \mid x ; W\right) \\
			& =\prod_{i,j} ( \hat{y}_{ij} )^{\widetilde{y}_{ij}} (1-\hat{y}_{ij})^{1-\widetilde{y}_{ij}} ,
		\end{split}
	\end{equation}
	where $i\in {1,...,M}$ and $j\in {1,...,N}$ represent the indices of the sample points in the vertical and horizontal dimensions, respectively. $\hat{y}=h(x ; W)$ is the prediction of the network, which we learn with original UNet~\cite{ronneberger2015u}.
	The UNet structure used in the training process is illustrated in Fig.~\ref{fig:SPR}.

	\subsection{Learning and Inference}
	By taking Eq.~(\ref{equ3}) and Eq.~(\ref{equ4}) into Eq.~(\ref{equ2}) and logging it, we obtain:
	\begin{equation}
		\label{equ5}
		\begin{split}
			\log \mathcal{L}(\widetilde{y}, W)=-&\sum_{k=1}^N \frac{\left\|t_k-s_k\right\|}{\gamma}\\+
			\sum_{i,j} &[\widetilde{y}_{ij} \log \hat{y}_{ij} 
			 +\left(1\! -\! \widetilde{y}_{ij}\right) \log (1-\hat{y}_{ij})].
		\end{split}
	\end{equation}
	
	To maximize the log likelihood function, we minimize its negative, $-\log \mathcal{L}(\widetilde{y}, W)$, using the strategy of updating $\widetilde{y}$ and $W$ alternately. By fixing $\widetilde{y}$ and optimizing $W$, the optimization is formulated as:
	\begin{equation}
		\label{equ:6}
		\min _W -\sum_{i,j} [\widetilde{y}_{ij} \log \hat{y}_{ij}+\left(1-\widetilde{y}_{ij}\right) \log (1-\hat{y}_{ij})].
	\end{equation}
	This is the binary cross-entropy function, which is often used in training networks for binary classification problems. The above equation means that we train the network with the current potential first-break $\widetilde{y}$ as the label. Note that our network learns the potential first-break $\widetilde{y}$ instead of the manual picking $y$. The update of $\widetilde{y}$ is obtained through the following optimization:
	\begin{equation}
		\label{equ:7}
		\min _{\widetilde{y} } \sum_{k=1}^N  \frac{\left\|t_k-s_k\right\|}{\gamma}-\sum_{i,j} [\widetilde{y}_{ij} \log \hat{y}_{ij}-\widetilde{y}_{ij} \log (1-\hat{y}_{ij})].
	\end{equation}
	Here, $\widetilde{y}$ and $s$ represent different forms of the same variable, making their optimization equivalent.
	The equation implies that updating $\widetilde{y}$ must align with the current network output and the artificial label $y$. The alternating updates of $\widetilde{y}$ and $W$ ensure the maximization of Eq.~(\ref{equ5}). The complete training process of SPR is shown in Algorithm 1.
	
	The trained model can perform inference in two ways: in the absence of $y$, it predicts the point of the first-break using a model trained to predict the potential true first-break more accurately than manual labeling. 
	The prediction is as follows:	
	\begin{equation}
        \label{eq:8}
		y^*=\underset{\widetilde{y}}{\arg \max }\  P(\widetilde{y} \mid x ; W).
	\end{equation}
	On the other hand, if manual picking $y$ is available, we can infer
	\begin{equation}
		\label{eq:9}
		y^{**}=\underset{\widetilde{y}}{\arg \max }\  P(y \mid \widetilde{y}) P(\widetilde{y} \mid x ; W)
	\end{equation}
	to refine manual picking, with $y^{**}$ being the refinement of $y$. The application of the two inferences is shown in subsequent experiments.
	
	\begin{algorithm}[t]
		\caption{SPR training Algorithm}
		\label{alg:moal}
		\begin{algorithmic}[1]
			
			\REQUIRE ~~\\ 
			Training set $\{x, y\}$, the number of epochs ${\emph{K}}$.
			\ENSURE ~~\\ 
			Network prediction $\hat{y}$.
			\STATE Initialize $W$, $\widetilde{y}=y$. Calculate the number of steps ${\emph{L}}$ based on the dataset size.
			\FOR {$epoch=1$ \text{\textbf{to}} ${\emph{K}}$}
			\FOR {$step=1$ \text{\textbf{to}} ${\emph{L}}$}
			\STATE Sample data $(x, y)$ and the corresponding $\widetilde{y}$.
			\STATE Predict $\hat{y}$ using current network parameters $W$.
			\STATE Update $W$ by Eq.~(\ref{equ:6}). 
			\STATE Update $\widetilde{y}$ by Eq.~(\ref{equ:7}).
			\ENDFOR
			\ENDFOR
		\end{algorithmic}
	\end{algorithm}
	
	\begin{table}[htbp]
		\huge
		\centering
		\caption{Information about the real data used in this paper. To adapt the input and output of the neural network, the original signal is filled with zeros to become the input shape.}
		\label{tab:data_infor}
		\resizebox{\linewidth}{!}{
			\begin{tabular}{cccccc}
				\toprule
				\multicolumn{6}{c}{\textbf{Data Information}} \\
				Dataset& trace num & sample num & sample rate & first-break & input shape \\ \midrule
				Sudbury~\cite{pierre2021multi} & 1810220 & 1001  & 1 ms  & 0-92 ms  & 224$\times$1024    \\
				Lalor~\cite{pierre2021multi}   &2424923&1501   & 2 ms  & 0-220 ms  & 192$\times$1504     \\
				
				\bottomrule
		\end{tabular}}
		
	\end{table}
	\section{Experiment}
	
		\begin{figure*}[!]
		\centering
		
		\includegraphics[width=\linewidth]{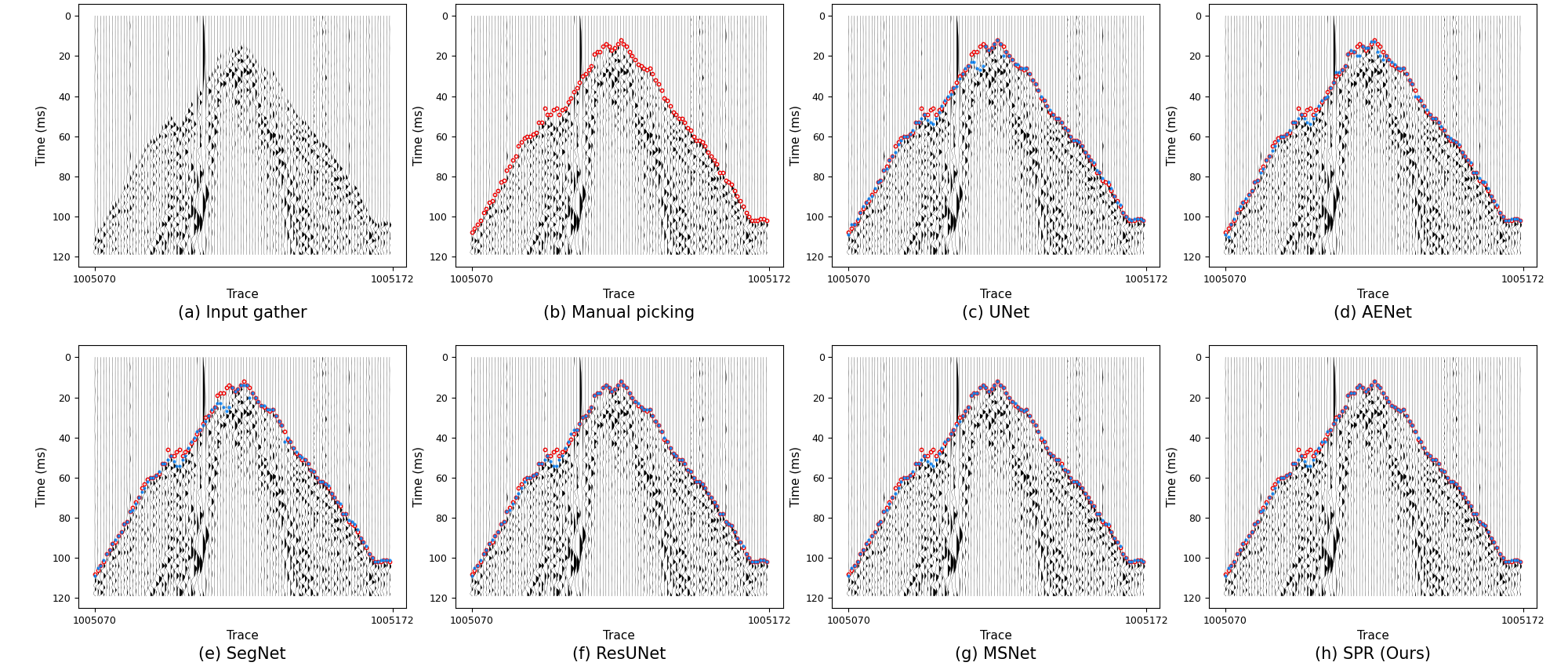}
		
		\caption{Comparison of picking results of different methods for trace 1005070-1005172 of dataset Sudbury. Manual picking and picking results of different methods are indicated by red circles and blue dots, respectively.}
		\label{fig:S_None}
	\end{figure*}
	
	\begin{figure*}[!]
		\centering
		
		\includegraphics[width=\linewidth]{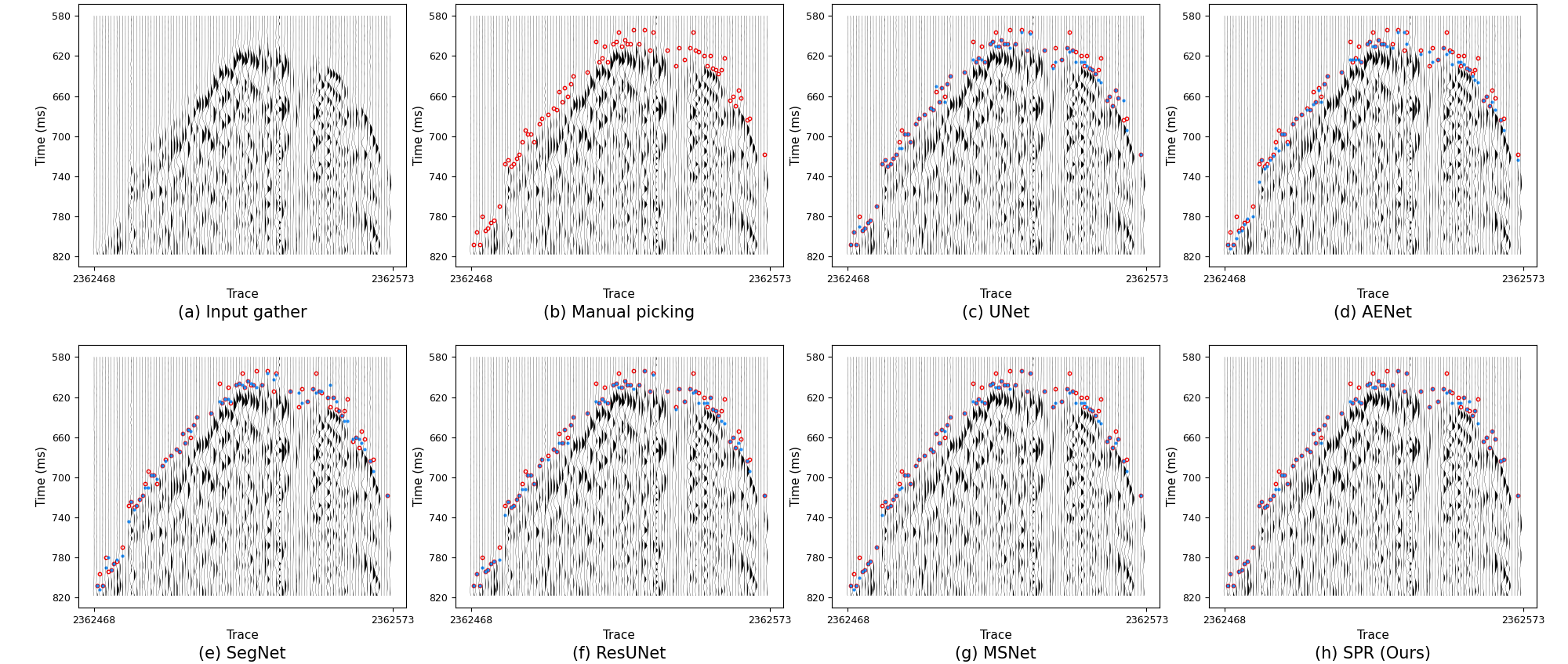}
		
		\caption{Comparison of picking results of different methods for trace 2362428-2362573 of dataset Lalor. Manual picking and picking results of different methods are indicated by red circles and blue dots, respectively.}
		\label{fig:L_None}
	\end{figure*}
	
	\subsection{Setup}

	All of our experiments are performed on two publicly available datasets. 
	The Sudbury dataset~\cite{pierre2021multi} contains 1,810,220 traces, each containing 1001 sampling points at a sampling rate of 1 ms.
	The Lalor dataset~\cite{pierre2021multi} contains 2,424,923 traces with 1001 sampling points and a sampling rate of 2 ms. 
	Basic details of the datasets are presented in Table~\ref{tab:data_infor}, with further information available in \cite{pierre2021multi}.
	In adapting to the network structure, the amplitude signal is processed to a size that can accommodate downsampling and upsampling, as shown in the last column of Tab.~\ref{tab:data_infor}, with the expanded region filled by zeros.
	Both datasets are divided into training set/validation set/test set according to the ratio of 8:1:1 for experimentation.
	Not all traces in the datasets are manually labeled, hence the picking results of different methods for these traces are omitted from the visualization.
	
	The experiment consists of four parts. The first part, a picking comparison experiment, is conducted on the Sudbury and Lalor datasets to compare the picking performance of different methods.
	The second part is a generalization experiment, where the models from the first part are directly transferred to assess the algorithm's generalization performance.
	The third part involves experiments with noisy signals, where Gaussian noise is artificially added to the Sudbury dataset and picking is performed under various noise levels.
	The last part involves experiments with noisy labels, conducted on the Sudbury dataset. 
    In the first three parts, the experimental results are inferred solely based on Eq.~(\ref{eq:8}). For the experiments with noisy labels, we apply Eq.~(\ref{eq:8}) on the test set to verify the ability to learn from noisy labels, and also apply Eq.~(\ref{eq:9}) on its training set to  confirm the capability to refine manual picking.
    
	The comparison methods in the experiments are comprised of the following state-of-the-art deep learning-based picking methods: UNet~\cite{hu2019first}, AENet~\cite{guo2020aenet}, SegNet~\cite{yuan2022segnet}, ResUNet~\cite{zwartjes2022first} and MSNet~\cite{sheng2022arrival}.
	The batch size, learning rate and optimizer are set to 8, 1e-4 and Adam, respectively.
	Parameter $\gamma$ is set to 5 to balance network output and refinement performance.
	The compared methods retain their original structures and are retrained using the same training settings employed for our SPR method.
	All models are trained for 100 epochs on a PC equipped with a single NVIDIA GeForce RTX 3090 GPU.
	\subsection{Evaluation}
	
	Two metrics are used to verify the accuracy and reliability of our method. 
	We transform the prediction result $y^*$ into 1-D form $t^*$ and use the 1-D manual picking $t$ as a reference.
	The first evaluation metric is hit rate (HR)~\cite{pierre2021multi}, which is defined as follows:
	
	\begin{equation}
		H R_\delta=\frac{1}{N} \sum_{k=1}^N I{\left(\left|t_k-t^*_k\right| \leq \delta\right)},
	\end{equation}
    where $N$ is the total number of channels, $I$ is the indicator function, and $\delta$ is the error parameter.
	HR reflects the proportion of correctly identified microseismic wave first-break out of all the potential seismic events present in a dataset.
	A high HR indicates a strong ability of the algorithm to accurately detect the first-break, which is crucial for subsequent seismic data processing tasks such as travel-time analysis.
	Additionally, the HR is instrumental in assessing the robustness of a picking method against various challenges, including noise levels, signal distortions, and variations in seismic wave characteristics.
	In this paper, we compare the performance of the algorithms at $\delta = \{0, 1, 2, 3, 5\}$.
	
	The second metric is the mean absolute error (MAE), which quantifies the average absolute difference between automatic picking and manual picking. 
	MAE provides a clear and interpretable measure of the average error magnitude, and it is highly resistant to outliers.
	MAE is defined as:
	\begin{equation}
	MAE=\frac{1}{N} \sum_{k=1}^N \left|t_k-t^*_k\right|.
	\end{equation}
	In the noisy label refining experiments, performance evaluation follows the same methodology as in picking performance. However, $t^*$ is replaced with $t^{**}$, which is derived from $y^{**}$.

	\begin{figure*}[!]
	\centering
	
	\includegraphics[width=\linewidth]{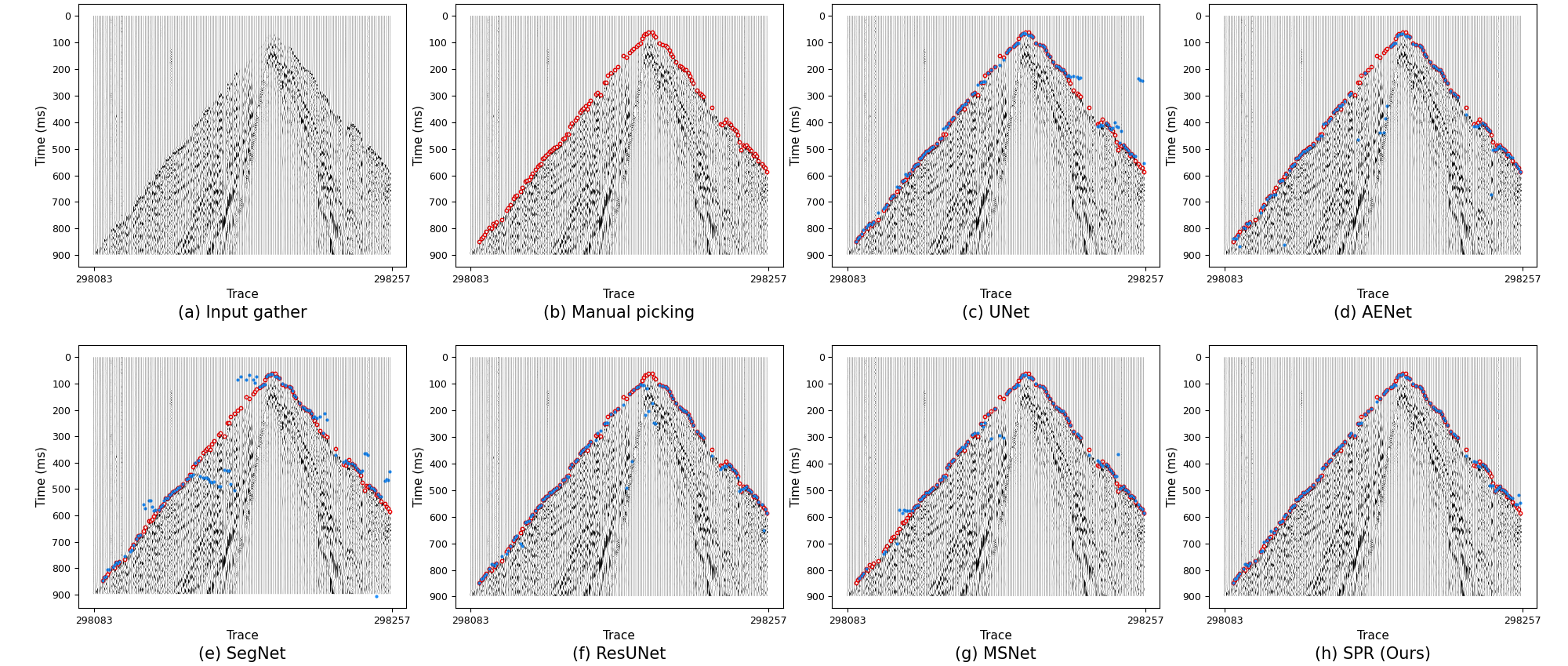}
	\caption{Generalization experiments for first-break picking. The model is trained on data Sudbury and tested on data Lalor. The figure shows traces 298083-298257 in data Lalor. Manual picking and generalization results of different methods are indicated by red circles and blue dots, respectively.}
	\label{fig:SL_None}
\end{figure*}

		\begin{table}[htbp]
	\centering
	\caption{Picking experiments for first-break picking. \textbf{Boldface} and \underline{underline} show the best and second best values, respectively.}
	\label{tab:None}
	\resizebox{\linewidth}{!}{
		\begin{tabular}{ccccccc}
			\toprule
			\multicolumn{7}{c}{\textbf{Dataset: Sudbury}} \\
			Method & $HR_0$ & $HR_1$ & $HR_2$ & $HR_3$ & $HR_5$ & $MAE$ \\ \midrule
			UNet~\cite{hu2019first}   & 65.31  & 89.10  & 91.56  & 93.06  & 94.91    & 2.6465  \\
			AENet~\cite{guo2020aenet}  &58.04 & 85.33 & 88.96  & 90.35  & 92.89 &    5.4552    \\
			SegNet~\cite{yuan2022segnet}& 64.92  & 88.84  & 91.35   & 92.86  & 94.68  & 2.3475  \\
			ResUNet~\cite{zwartjes2022first}& 68.11  & 89.34  & 91.47 & 92.86  & 94.55   & 3.5663  \\
			MSNet~\cite{sheng2022arrival}& \underline{69.03}  & \underline{89.85}  & \underline{92.16} & \underline{93.59} & \underline{95.25}   & \underline{2.2113}  \\
			SPR (Ours)  & \textbf{74.20} & \textbf{90.26} & \textbf{92.35} & \textbf{93.75} & \textbf{95.63} & \textbf{1.6014} \\ 
			\midrule
			\multicolumn{7}{c}{\textbf{Dataset: Lalor}} \\
			Method & $HR_0$ & $HR_1$ & $HR_2$ & $HR_3$ & $HR_5$ & $MAE$ \\ \midrule
			UNet~\cite{hu2019first}    & 90.28  & 92.15  & 92.80  & 93.72  & 95.43    & 0.5305  \\
			AENet~\cite{guo2020aenet}   &  71.97 &   86.51& 88.12  & 89.50 & 92.75    &  1.2198 \\
			SegNet~\cite{yuan2022segnet}& 76.98  & 89.11  & 90.69   & 92.09  & 94.50  & 0.7988  \\
			ResUNet~\cite{zwartjes2022first}& 91.20  & 92.09  & 92.72 & 93.58  & \underline{95.56}   & 0.5375  \\
			MSNet~\cite{sheng2022arrival}& \underline{91.91}  & \underline{92.56}  & \underline{93.10} & \underline{93.87} & 95.52   & \underline{0.5129}  \\
			SPR (Ours)  & \textbf{92.60} & \textbf{92.80} & \textbf{93.34} & \textbf{94.14} & \textbf{95.76} & \textbf{0.4821} \\ 
			\bottomrule
			
	\end{tabular}}
\end{table}%
	\subsection{Picking Comparison Experiment}
	\label{picking_compare}
	We compare the picking performance of SPR and other methods on dataset Sudbury and dataset Lalor and present the quantitative results in Tab.~\ref{tab:None}.
	Two cases from dataset Sudbury and dataset Lalor are shown in Fig.~\ref{fig:S_None} and Fig.~\ref{fig:L_None}, respectively, where manual picking are represented by red circles and automatic picking are represented by blue dots.
	
	SPR exhibits a higher HR and a lower MSE, in particular, $HR_0$ is 5$\%$ higher than the suboptimal value, suggesting that our method is more accurate and reliable in correctly identifying the first-break.
	In the visualization results, there are traces for which manual picking positions cannot be identified by all methods, as exemplified by the 40-60 ms interval in Fig.~\ref{fig:S_None}.
	Raw learning of such traces has the potential to degrade the algorithm's performance on general traces. In contrast, SPR's learning targets are not based on manual pickings and are therefore unaffected by outlier samples, ensuring more consistent picking performance across the dataset.
	
	\subsection{Picking Generalization Experiments}
	\begin{table}[htbp]
		\centering
		\caption{Generalization experiments for first-break picking. \textbf{Boldface} and \underline{underline} show the best and second best values, respectively.}
		\label{tab:SL}
		\resizebox{\linewidth}{!}{
			\begin{tabular}{ccccccc}
				\toprule
				\multicolumn{7}{c}{\textbf{Training: Sudbury~~~~~ Test: Lalor}} \\
				Method & $HR_0$ & $HR_1$ & $HR_2$ & $HR_3$ & $HR_5$ & $MAE$ \\ \midrule
				UNet~\cite{hu2019first}   & 36.34  & 58.83  & 66.11  & 71.32  & 79.02    & 15.8180  \\
				AENet~\cite{guo2020aenet}  &52.30 & 70.85 & 73.41  & 75.84  & \underline{81.97} &    9.2749    \\
				SegNet~\cite{yuan2022segnet}& 43.98  & 63.71  & 66.18   & 68.42  & 73.95  & 21.1314  \\
				ResUNet~\cite{zwartjes2022first}& \underline{56.14}  & \underline{73.80}  & \underline{75.22} & \underline{77.25}  & 81.95   & \textbf{5.1749}  \\
				MSNet~\cite{sheng2022arrival}& 54.00  & 72.06  & 74.36 & 76.66 & 81.88   & 14.7028  \\
				SPR (Ours)  & \textbf{58.44} & \textbf{76.62} & \textbf{78.90} & \textbf{80.96} & \textbf{85.63} & \underline{6.2842} \\ 
				\bottomrule

		\end{tabular}}
	\end{table}%

	Given the variability in seismic data due to geographic location, geological structure, and acquisition techniques, it is crucial for a first-break picking algorithm to demonstrate both efficiency and accuracy under these diverse conditions. To investigate the applicability and robustness of our picking algorithm, as well as its ability to generalize across different survey sites, we conducted generalization experiments on the Sudbury and Lalor datasets. 
    The rationale behind this setup lies in the significant difference between Sudbury and Lalor datasets, as presented in Table~\ref{tab:data_infor}. Original study~\cite{pierre2021multi} also indicates that knowledge generalization between them is quite challenging, making them suitable subjects for a generalization experiment.
    The models trained on the Sudbury dataset, as described in Sec.~\ref{picking_compare} are directly tested on the test set of Lalor dataset, and the results of the comparison are displayed in Tab.~\ref{tab:SL} and Fig.~\ref{fig:SL_None}.

	The comparison results show that, due to differences in location and sampling rates between the Sudbury (1 ms) and Lalor (2 ms) datasets, the performance of all methods experienced some degradation. However, our method is minimally impacted by these differences and maintained a high degree of consistency with manual picking. SPR demonstrated superior generalization performance compared to other algorithms that learn directly from artificial labels. This suggests that SPR effectively captures the true distribution of the data, thereby avoiding overfitting and enhancing its robustness to various challenges.

		\begin{figure*}[!]
		\centering
		
		\includegraphics[width=\linewidth]{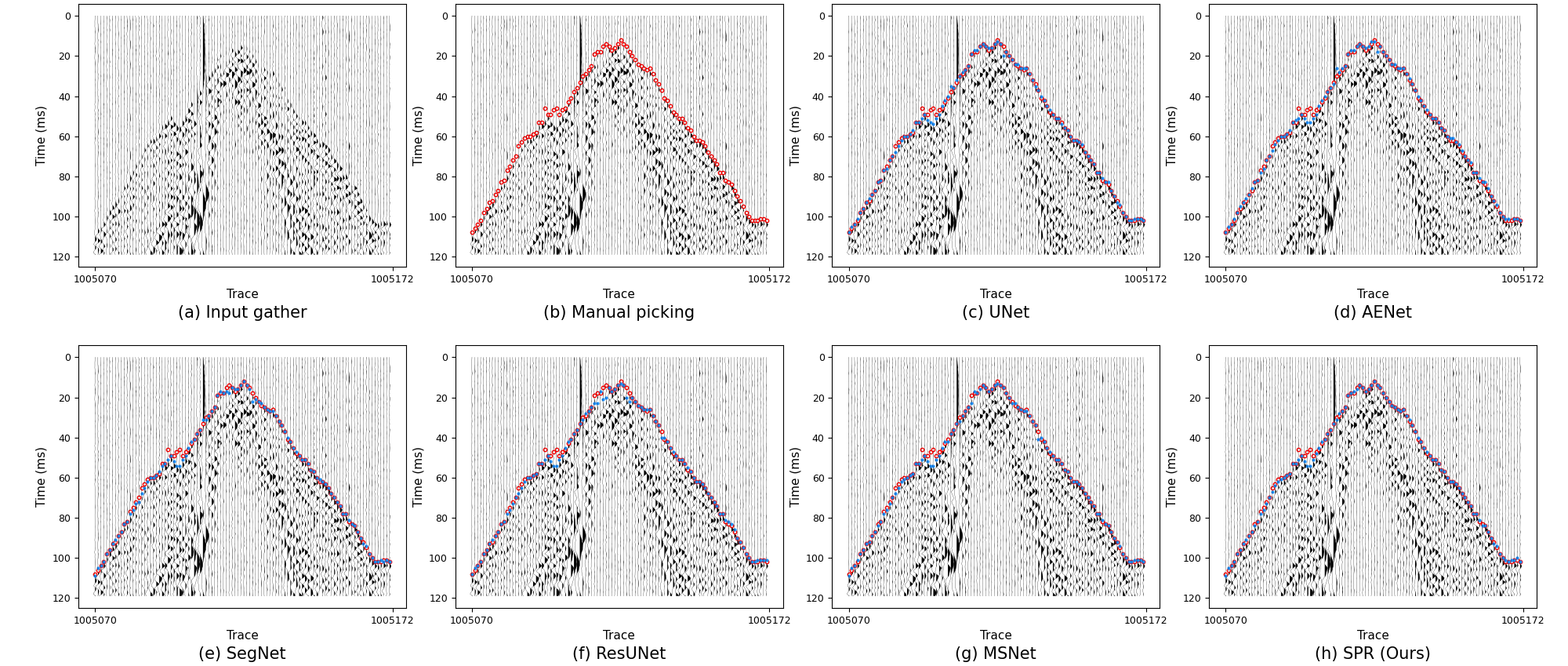}
		
		\caption{Comparison of picking results of different methods for trace 1005070-1005172 of dataset Sudbury at noise level 0.05. Manual picking and picking results of different methods are indicated by red circles and blue dots, respectively.}
		\label{fig:S_0.05}
	\end{figure*}

	\begin{figure*}[!]
		\centering
		
		\includegraphics[width=\linewidth]{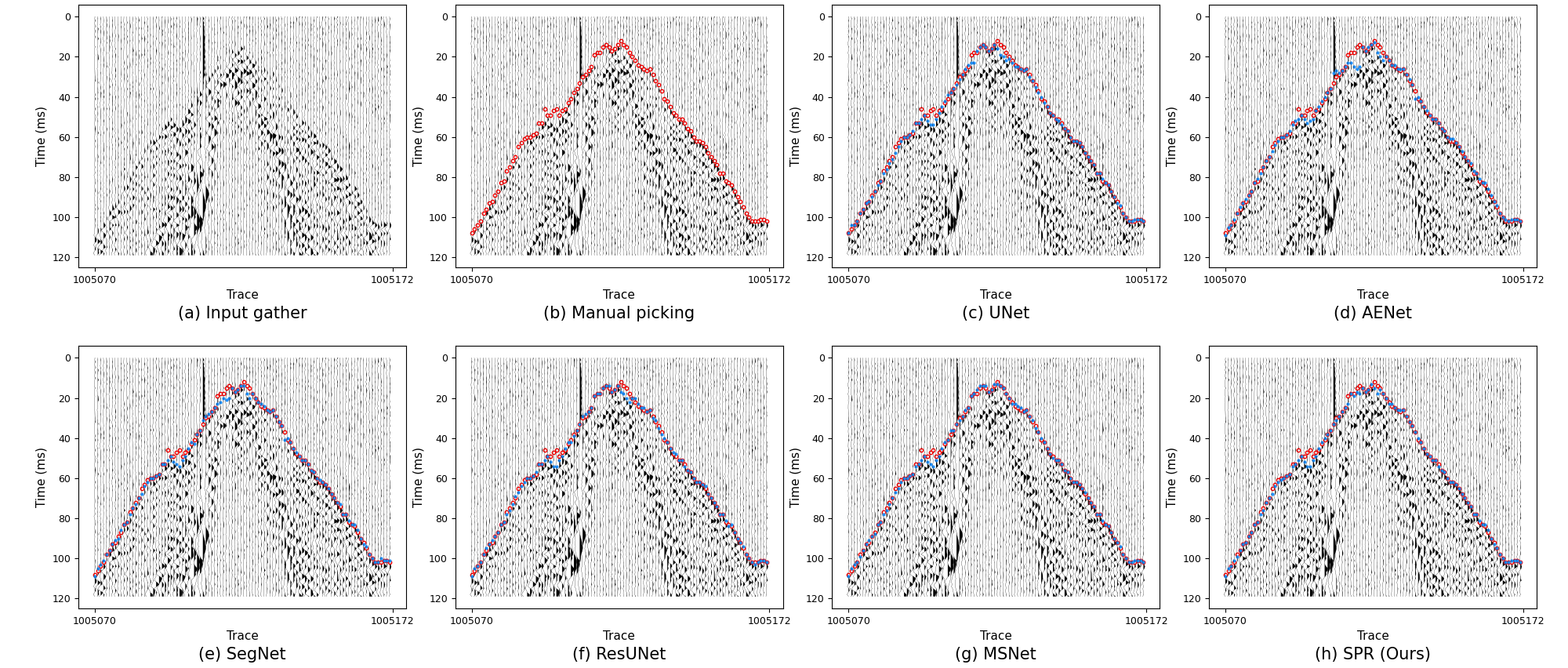}
		
		\caption{Comparison of picking results of different methods for trace 1005070-1005172 of dataset Sudbury at noise level 0.1. Manual picking and picking results of different methods are indicated by red circles and blue dots, respectively.}
		\label{fig:S_0.1}
	\end{figure*}
	
	\begin{figure*}[!]
		\centering
		
		\includegraphics[width=\linewidth]{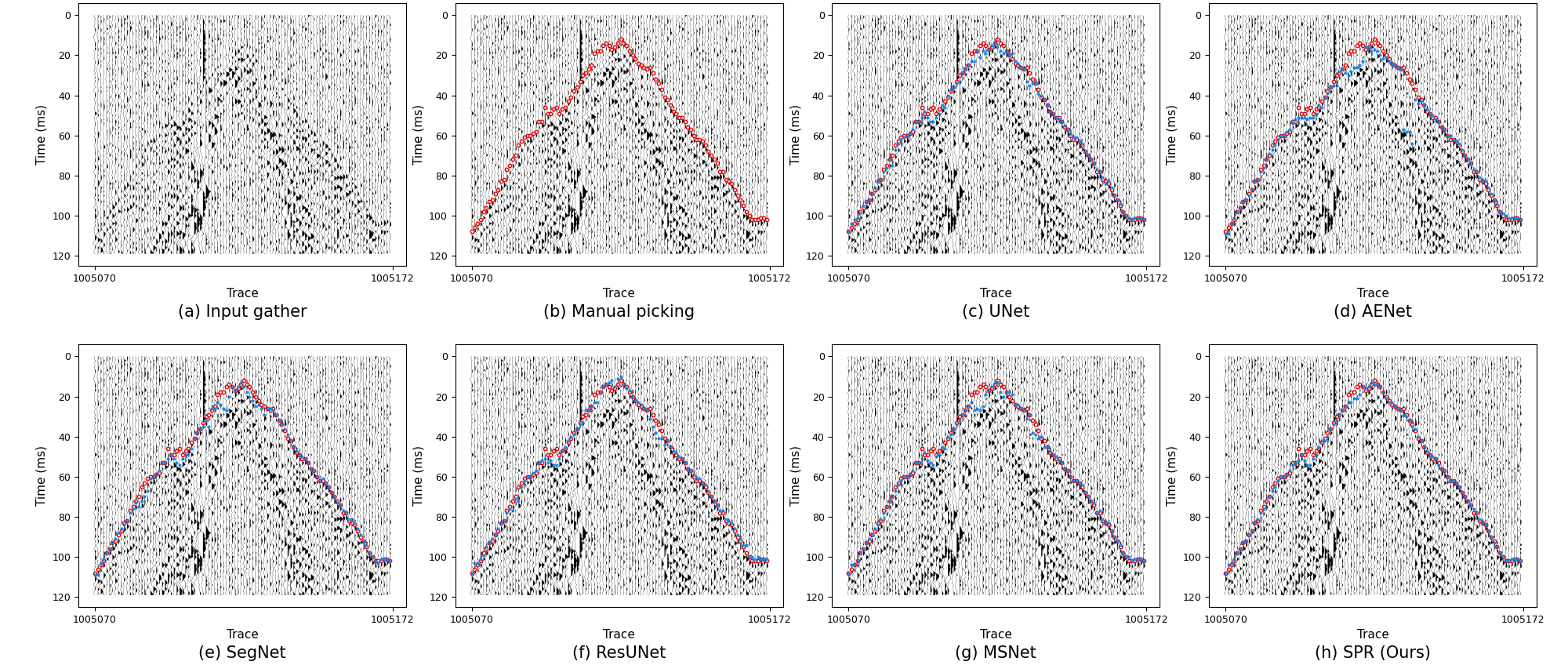}
		
		\caption{Comparison of picking results of different methods for trace 1005070-1005172 of dataset Sudbury at noise level 0.2. Manual picking and picking results of different methods are indicated by red circles and blue dots, respectively.}
		\label{fig:S_0.2}
	\end{figure*}
	\subsection{Noisy Signal Experiment}
	\label{Noisy Signal Experiment}
	In this section, Gaussian noise with a mean of 0 is added to the signal to study the noise immunity of the picking method at different noise levels.
	Following~\cite{kim2023first}, the standard deviation of the Gaussian noise added to each trace is determined by multiplying the trace's maximum raw data amplitude by the noise level (NL). The standard deviation of the noise $\sigma_n$ in the $k$-th trace is calculated as follows:
	 \begin{equation}
		\sigma_n=max(abs(x_k))* \textit{NL},
	 \end{equation}
	where $x_k$ is the signal of the $k$-th trace.
	We choose three noise levels: 0.05, 0.1, and 0.2, to verify the picking performance on the Sudbury dataset.
	Comparative results of the same example in Sec.~\ref{picking_compare} at different noise levels are shown in Fig.~\ref{fig:S_0.05}, Fig.~\ref{fig:S_0.1}, and Fig.~\ref{fig:S_0.2}. 
	With increasing noise levels, the amplitude signal is gradually drowned out by noise, and all methods are affected to varying degrees. Compared to the other methods, SPR is still able to obtain less biased picking results under strong noise conditions.
	The quantitative results in Table~\ref{tab:Noisy Signal} demonstrate that our method has the highest HR and lowest MAE at all noise levels, showcasing the great robustness of SPR in heavily noisy environments.
	
	\begin{table}[htbp]
		\centering
		\caption{Noisy signal experiments for first-break picking. \textbf{Boldface} and \underline{underline} show the best and second best values, respectively.}
		\label{tab:Noisy Signal}
		\resizebox{\linewidth}{!}{
			\begin{tabular}{ccccccc}
				\toprule
				\multicolumn{7}{c}{\textbf{Dataset: Sudbury~~~~~Noise Level = 0.05}} \\
				Method & $HR_0$ & $HR_1$ & $HR_2$ & $HR_3$ & $HR_5$ & $MAE$ \\ \midrule
				UNet~\cite{hu2019first}   & 60.45  & 87.46  & 90.20  & 91.87  & 93.97    & 4.9023  \\
				AENet~\cite{guo2020aenet}    &   54.79&83.39   &87.16   & 89.00 & 91.05    & 9.3147  \\
				SegNet~\cite{yuan2022segnet} & 60.44  & 87.45  & \underline{90.78}   & \underline{92.65}  & \underline{94.95}  & \underline{1.9859}  \\
				ResUNet~\cite{zwartjes2022first}& 62.18  & \underline{87.77}  & 90.66 & 92.20  & 94.20   & 2.9811  \\
				MSNet~\cite{sheng2022arrival}& \underline{62.48}  & 87.51  & 90.52 & 92.25  & 94.29   & 3.5436  \\
				SPR (Ours)  & \textbf{65.43} & \textbf{88.68} & \textbf{91.67} & \textbf{93.41} & \textbf{95.39} & \textbf{1.7251} \\ 
				\midrule
				\multicolumn{7}{c}{\textbf{Dataset: Sudbury~~~~~Noise Level = 0.1}} \\
				Method & $HR_0$ & $HR_1$ & $HR_2$ & $HR_3$ & $HR_5$ & $MAE$ \\ \midrule
				UNet~\cite{hu2019first}    & 57.49  & 85.86  & 90.02  & 91.96  & 94.41    & 2.5230  \\
				AENet~\cite{guo2020aenet}    &  47.77 & 79.87  & 85.16  & 87.64 &  90.24   & 10.1634  \\
				SegNet~\cite{yuan2022segnet} & 53.39  & 84.16  & 88.91   & 91.00  & 93.51  & 3.1368  \\
				ResUNet~\cite{zwartjes2022first}& 58.06  & \underline{86.40}  & \underline{90.17} & 92.06  & 94.27   & 2.8152  \\
				MSNet~\cite{sheng2022arrival}& \underline{58.13}  & 86.30  & 90.14 & \underline{92.16}  & \underline{94.41}   & \underline{2.4048}  \\
				SPR (Ours)  & \textbf{59.71} & \textbf{87.20} & \textbf{91.14} & \textbf{93.11} & \textbf{95.25} & \textbf{1.8380} \\ 
				\midrule
				\multicolumn{7}{c}{\textbf{Dataset: Sudbury~~~~~Noise Level = 0.2}} \\
				Method & $HR_0$ & $HR_1$ & $HR_2$ & $HR_3$ & $HR_5$ & $MAE$ \\ \midrule
				UNet~\cite{hu2019first}    & 46.45  & 78.95  & 84.67  & 87.55  & 90.95    & 4.5489  \\
				AENet~\cite{guo2020aenet}   &37.68 &69.30   & 76.67  & 80.04  & 84.03 &   11.1791     \\
				SegNet~\cite{yuan2022segnet} & 44.79  & 77.55  & 84.43   & 87.46  & 91.07  & 3.7183  \\
				ResUNet~\cite{zwartjes2022first}& 46.66  & \underline{79.84}  & 85.59 & 88.28  & \underline{91.65}   & 3.7779  \\
				MSNet~\cite{sheng2022arrival}& \underline{47.53}  & 79.53  & \underline{85.70} & \underline{88.38}  & 91.65   & \underline{3.7066}  \\
				SPR (Ours)  & \textbf{50.06} & \textbf{83.07} & \textbf{88.81} & \textbf{91.64} & \textbf{94.67} & \textbf{1.9638} \\ 
				
				\bottomrule
				
		\end{tabular}}
	\end{table}%
	
	\begin{table}[htbp]
		\centering
		\caption{Performance of noisy labels experiments. \textbf{Boldface} and \underline{underline} show the best and second best values, respectively.}
		\label{tab:nl_train}
		\resizebox{\linewidth}{!}{
			\begin{tabular}{ccccccc}
				\toprule
				\multicolumn{7}{c}{\textbf{Label Refinement Performance of Noisy Label Experiments}} \\
				Methods & $HR_0$ & $HR_1$ & $HR_2$ & $HR_3$ & $HR_5$ & $MAE$ \\ \midrule
				Noisy Label & 13.30& 38.35  & 59.44  & 75.62  & 93.31   & 2.4134  \\ \midrule
				UNet~\cite{hu2019first}   &43.97&83.59   & 90.18  & 92.41  & 94.26 &   3.3341     \\
				AENet~\cite{guo2020aenet}    &28.42&67.94   & 82.66 & 87.36  & 90.41 &   9.8110     \\
				
				SegNet~\cite{yuan2022segnet} & 42.49 & 83.22  & 90.92   & 92.23  & \underline{95.29} & 2.8869  \\
				
				ResUNet~\cite{zwartjes2022first}& \underline{46.79}  & \underline{85.15}  & 90.78& 92.76  & 94.53   & 3.0953 \\
				MSNet~\cite{sheng2022arrival}& 43.55 & 84.84  & \underline{91.54} & \underline{93.41}  & 95.01  & \underline{2.8739} \\
				SPR (Ours)  & \textbf{56.24} & \textbf{88.75} & \textbf{92.60} & \textbf{94.20} & \textbf{95.79} & \textbf{1.7347} \\ 
				\midrule
				\multicolumn{7}{c}{\textbf{Picking Performance of Noisy Label Experiments}} \\
				Methods & $HR_0$ & $HR_1$ & $HR_2$ & $HR_3$ & $HR_5$ & $MAE$ \\ \midrule
				UNet~\cite{hu2019first}   & 44.53  & 83.47  & 90.95  & 93.22  & 95.10    & 2.6880 \\
				AENet~\cite{guo2020aenet}   &28.59&68.40   & 83.34  & 88.18  & 91.35 &   9.1939     \\
				SegNet~\cite{yuan2022segnet} & 42.76 & 82.94  & 90.71   & 93.24  & 95.18  & 2.3577  \\
				ResUNet~\cite{zwartjes2022first}& 45.10  & \underline{84.64}  & 91.48& 93.65  & \underline{95.36}   & 2.4516  \\
				MSNet~\cite{sheng2022arrival}& \underline{45.29} & 84.53  & \underline{91.59} & \underline{93.68}  & 95.23   & \underline{2.3580} \\
				SPR (Ours)  & \textbf{55.97} & \textbf{88.45} & \textbf{92.26} & \textbf{93.92} & \textbf{95.61} & \textbf{2.0987} \\ 
				\bottomrule
				
		\end{tabular}}
	\end{table}%
	\subsection{Noisy Label Experiment}\label{Sec:Noisy Label Experiment}
	
		\begin{figure*}[!]
		\centering
		\includegraphics[width=\linewidth]{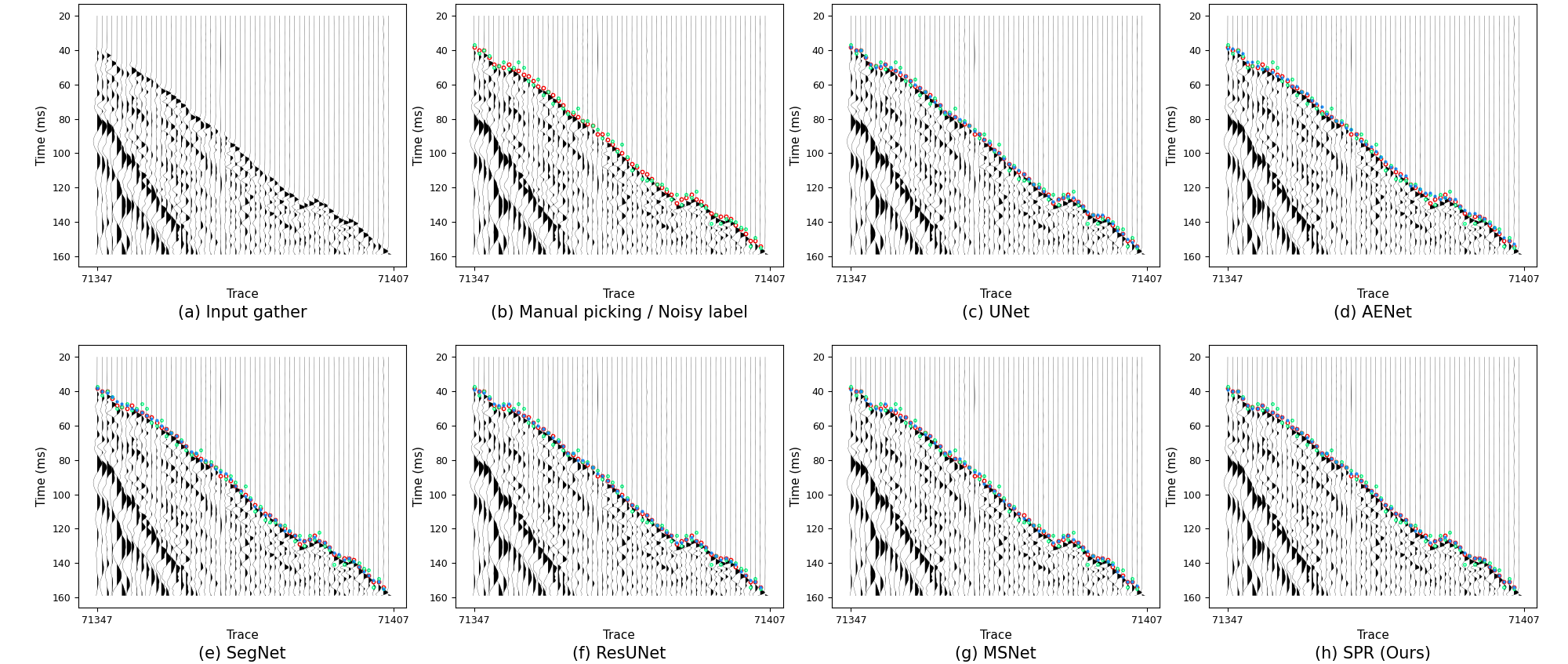}
		
		\caption{Comparison of noisy label refinement of different methods for trace 71347-71407 of dataset Sudbury. Manual picking and picking results of different methods are indicated by red circles and blue dots, respectively. Noisy labels are indicated by green circles.}
		\label{fig:S_nl_train}
	\end{figure*}
	
	\begin{figure*}[!]
		\centering
		\includegraphics[width=\linewidth]{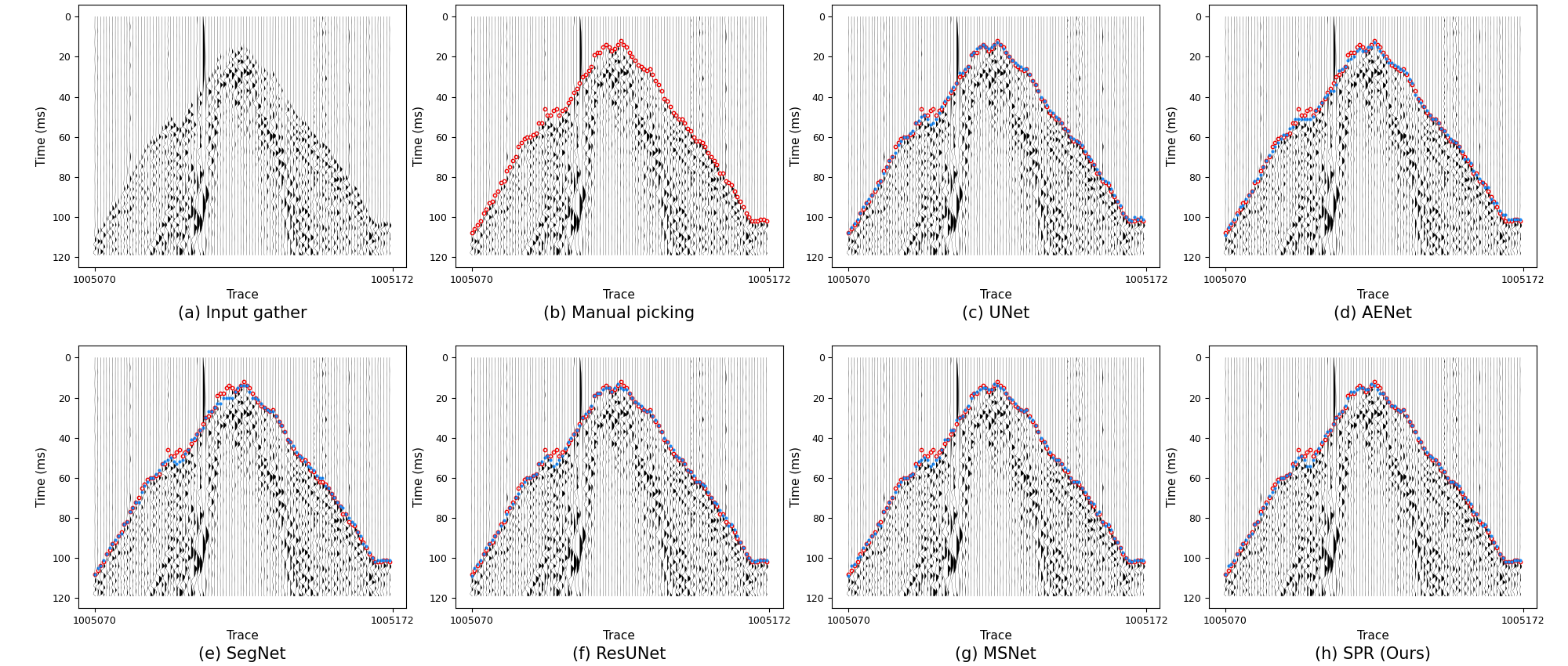}
		\caption{Comparison of noisy label picking of different methods for trace 1005070-1005172 of dataset Sudbury. Manual picking and picking results of different methods are indicated by red circles and blue dots, respectively.}
		\label{fig:S_nl_test}
	\end{figure*}

	What distinguishes our approach from others is the ability to learn correctly from undesirable labeled data.
	The labeling of the original data is always considered clean and correct. 
	In this section, we add labeling noise to all the training set labels of the Sudbury dataset to validate the automatic picking method's performance in learning from noisily labeled data.
	We model the distribution of mislabeling for each trace with a Gaussian distribution, the noisy labels are calculated as follows,
	 \begin{equation}
		\ddot{t}_k=t_k+\epsilon,
	 \end{equation}
	where $t_k$ is the indexed manual picking of the $k$-th trace, $\ddot{t}_k$ is the noisy labeling after noise addition, $\epsilon$ is Gaussian noise with mean 0 and variance 3.
	The addition of noise causes the first-break label of the training set to be shifted around the time dimension, and the noisy labels are marked with green circles in Fig.~\ref{fig:S_nl_train}.
	The first row of Table~\ref{tab:nl_train} reveals the deviation between the noisy labels and the original manual picking, showing an average deviation of 2.4 sampling points. Only 13.30\% of the labels remain in their original positions, indicating significant displacement for the majority.
	
	The training set accompanied by noisy labels is used to train the neural network. For SPR, we utilized artificial labels and employed the results of Eq.~(\ref{eq:9}) as the refined labels.
	For the other compared methods, we directly use their output as the refinement results. The results of the label refinement are displayed in Tab.~\ref{tab:nl_train} and Fig.~\ref{fig:S_nl_train}.
	Due to the degradation of label quality, no method other than SPR is able to achieve effective picking. SPR withstands the noise and can successfully improve the HR$_0$ of label refinement to 56.24\%, nearly 10\% higher than the suboptimal value, despite only 13.30\% of the labels being in the correct position.
	Furthermore, our method uniquely refines MAE below that of the noisy labels, indicating strong tolerance to label noise. 
	In contrast, other methods show more biased refinement results influenced by the noisy labels.
	The visualization in Fig.~\ref{fig:S_nl_train} also demonstrates that our refinement results are more consistent with the true first-break whereas other methods fail to obtain satisfactory results.
	
	Second, the model trained on the training set with noisy labels is applied to the test set of the Sudbury dataset to verify the picking performance.
	Manual picking on the test set is only used to verify picking accuracy, therefore, the green circles with noise labels are not represented in Fig.~\ref{fig:S_nl_test}. The samples selected remain the same as in Sections \ref{picking_compare} and \ref{Noisy Signal Experiment}.
	It is evident that the picking performance of the compared methods, as shown in the middle of Fig.~\ref{fig:S_nl_test}, is significantly impaired by the presence of noise in the labels, whereas our method continues to guarantee excellent picking results.
	Quantitative comparisons, as shown in Tab.~\ref{tab:nl_train}, reveal that our method achieves superior HR across all $\delta$ values, along with the lowest MAE. This success is attributed to our modeling and learning strategy for the potential first-break.
	
	\begin{table}[htbp]
		\centering
		\caption{Parameter investigation for first-break picking. \textbf{Boldface} shows the best values.}
		\label{tab:para}
		\resizebox{0.9\linewidth}{!}{
			\begin{tabular}{ccccccc}
				\toprule
				\multicolumn{7}{c}{\textbf{Parameter Investigation: Picking Experiments}} \\
				$\gamma$ & $HR_0$ & $HR_1$ & $HR_2$ & $HR_3$ & $HR_5$ & $MAE$ \\ \midrule
				500   & 55.11  & 79.69  & 85.43  & 87.63  & 93.57    & 6.4912  \\
				50    &   60.18&84.61   &88.19 & 89.16& 93.99   & 5.1945 \\
				5 & \textbf{74.20}  & \textbf{90.26}  & \textbf{92.35}   & \textbf{93.75}  & \underline{95.63}  & \textbf{1.6014}  \\
				0.5& \underline{73.26}  & \underline{90.18}  & \underline{92.30} & \underline{93.74}  & \textbf{95.64}  & \underline{1.7769}  \\
				0.05 & 67.22  & 89.53  & 91.41 & 93.28  & 95.39   & 2.3342  \\
				
				\midrule
				\multicolumn{7}{c}{\textbf{Parameter Investigation: Noisy Label Refinement}} \\
				$\gamma$ & $HR_0$ & $HR_1$ & $HR_2$ & $HR_3$ & $HR_5$ & $MAE$ \\ \midrule
				500 & 40.69  & 79.49  & 87.95  	& 90.33  & 91.19   & 5.9617  \\
				50  & 44.69  & 83.13  &90.25 	& 92.56  & 92.68   & 4.6519  \\
				5 	& \textbf{56.24}  & \textbf{88.75}  & \textbf{92.60} 	& \textbf{94.20}  & \textbf{95.79}   & \textbf{1.7347}  \\
				0.5	& \underline{50.26}  & \underline{86.99}  & \underline{92.02}	& \underline{94.13}  & \underline{95.51}   & \underline{2.8361}  \\
				0.05& 46.83  & 85.24  & 91.89 	& 93.96  & 95.49   & 2.9477  \\ 

				\midrule
				\multicolumn{7}{c}{\textbf{Parameter Investigation: Noisy Label Picking}} \\
				$\gamma$ & $HR_0$ & $HR_1$ & $HR_2$ & $HR_3$ & $HR_5$ & $MAE$ \\ \midrule
				500   & 25.97  & 60.19  & 75.49  & 83.49  & 90.12    & 9.3734  \\
				50    &   40.26&75.18   &83.94   & 88.64 & 89.17    & 7.1682  \\
				5 & \textbf{55.97}  & \textbf{88.45}  & \textbf{92.26}   & \textbf{93.92}  & \textbf{95.61}  & \textbf{2.0987}  \\
				0.5& \underline{54.68}  & \underline{87.49}  &\underline{91.98} & \underline{93.81}  & \underline{95.32}   & \underline{2.1653}  \\
				0.05 & 44.85  & 84.99  & 91.46 & 93.70  & 95.33   & 2.4397  \\ 
				
				\bottomrule
				
		\end{tabular}}
	\end{table}%
	\subsection{Parameter Investigation}
	In this section, we investigate the impact of parameter $\gamma$ on the performance. Picking experiments and experiments with noisy labels are performed on the Sudbury dataset for different $\gamma$ values, while the rest of the training setup parameters remain unchanged. The results of the investigation are displayed in Tab.~\ref{tab:para}. 
	
	When $\gamma$ is too small, the first term in Eq.~(\ref{equ:7}) dominates, and the optimization result of $\widetilde{y}$ is highly consistent with $y$.
	Therefore, the algorithm will gradually degenerate into an algorithm with manual picking as the learning objective, losing the ability to mine potential first-break. Consequently, there is a slight decrease in both the picking and correcting abilities, as shown in Tab.~\ref{tab:para}.
	When $\gamma$ is too large, the network's learning target heavily relies on updating $\widetilde{y}$ in the early training stages, which impacts the training outcomes and proves even less effective than directly learning manual picking.
	Therefore, in this paper, we choose $\gamma = 5$ to balance the learning of the network as well as the mining of potential first-break.
	
\section{Discussion and Conclusion}
\subsection{Discussion}
	In our formulation of Eq.~(\ref{equ2}), we posit that the manual labeling $y$ is independent of $x$ given the true first-break $\tilde{y}$, allowing for the simplification of $P(y \mid \widetilde{y}, x ; W)$ to $P(y \mid \widetilde{y})$. This assumption is premised on the notion that errors in manual labeling are predominantly influenced by incomplete accuracy of the labeling (imprecision around the true first-break point) rather than by errors arising from external factors like a lack of expertise (marking one period too early or too late), differing annotation standards (choosing among peaks, troughs, or onset points), or the characteristics of $x$ itself. This modeling strategy also effectively separates the network's predictive terms from the prior labeling terms within the likelihood function, streamlining both analysis and computation.

The Laplace distribution, chosen for its sharp peak and ease of computation, particularly in log-likelihood calculations, is apt for modeling labeling errors due to its higher probability density near zero compared to Gaussian and $t$-distributions. Its frequent application in prior distribution modeling further motivated its selection in our study for capturing potential labeling inaccuracies.

However, the efficacy of the Laplace distribution heavily depends on the scale parameter. Our approach employs the strategy of a uniform scale parameter. 
This strategy simplifies the model but it overlooks variability in sample sizes and unknown standard deviations, which can be addressed more effectively by more flexible distributions like the $t$-distribution. 
This limitation suggests that for more complex signal types or more diverse error sources, a uniform scale parameter may not be universally optimal. Moreover, the fixed statistical properties of the Laplace distribution, such as its skewness and kurtosis, could restrict its applicability in situations demanding greater statistical flexibility.

Looking ahead, we aim to explore more adaptable statistical models, including the $t$-distribution and asymmetric distributions, which offer better accommodation for varied sample sizes and complex signal characteristics. Additionally, we plan to investigate the potential benefits of modeling $P(y \mid \widetilde{y}, x ; W)$ in relation to $x$, to enhance our model's adaptability to the complex scenarios presented by $x$. This future work will strive to address the limitations identified and to refine our approach for even more accurate and flexible seismic data analysis.

\subsection{Conclusion}
In this article, we propose a novel first-break picking algorithm targeting challenges posed by outlier samples and noisy labels.
Our method constructs a joint probabilistic model that combines manual picking with potential true first-break. These are treated as variables and are alternately optimized alongside neural network parameters.
The core innovation of our algorithm is its dual capability. 
It can execute first-break picking with enhanced precision and refine the manual picking. 
This dual functionality is validated by extensive experiments on real seismic data, including comparisons of picking performance, generalization capabilities, and tests on noisy signals and noisy labels.
The results consistently underscore the robustness and applicability of our approach in accurately identifying first-break in seismic data.
Furthermore, our algorithm provides a modeling approach for first-break picking, applicable to both training and inference, that can be adapted to neural networks of any structure.

\bibliographystyle{IEEEtran}
\bibliography{refer}
\end{document}